\documentclass[10pt,twocolumn,letterpaper]{article}

\usepackage{cvpr}              

\usepackage{cuted}

\definecolor{cvprblue}{rgb}{0.21,0.49,0.74}
\usepackage[pagebackref,breaklinks,colorlinks,allcolors=cvprblue]{hyperref}
\usepackage{todonotes, bm}

\def\paperID{16} 
\def\confName{CVPR}
\def\confYear{2025}

\title{Text2Stereo: Repurposing Stable Diffusion for Stereo Generation with Consistency Rewards}

\author{%
Aakash Garg\textsuperscript{1}
\quad Libing Zeng\textsuperscript{1}
\quad Andrii Tsarov\textsuperscript{2}
\quad Nima Khademi Kalantari\textsuperscript{1}\\
\textsuperscript{1}Texas A\&M University, \textsuperscript{2}Leia Inc.\\
{\tt\small \{aakash.garg80,libingzeng,nimak\}@tamu.edu, andrii.tsarov@leiainc.com}
}

\begin{document}
\maketitle


\begin{strip}\centering
\vspace{-0.5in}
\includegraphics[width=1.0\textwidth]{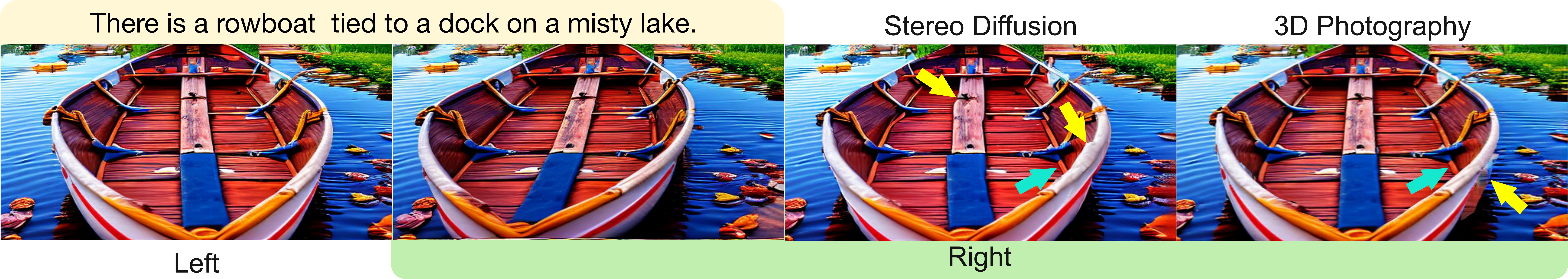}
\vspace{-0.25in}
\captionof{figure}{Given an input text prompt, our method synthesizes a stereo pair of left and right images. We use the generated left image as input to StereoDiffusion~\cite{wang2024stereodiffusion} and 3D Photography~\cite{3d_photography_2020} to generate the right image. Both StereoDiffusion and 3D Photography use depth-based warping to transfer content from the input to the novel view. As such, they often struggle to create appropriate parallax effects for objects with continuously varying depth, as indicated by the cyan arrows. Moreover, since StereoDiffusion performs depth warping in the latent space, the warping is often not pixel-perfect, resulting in objectionable artifacts, as indicated by the yellow arrows. Finally, 3D Photography frequently struggles to reconstruct occluded regions (see the yellow arrow). Our approach, however, produces consistent, high-quality stereo images with wide baselines}
\vspace{-0.1in}
\label{fig:teaser}
\end{strip}


\begin{abstract}
In this paper, we propose a novel diffusion-based approach to generate stereo images given a text prompt. Since stereo image datasets with large baselines are scarce, training a diffusion model from scratch is not feasible. Therefore, we propose leveraging the strong priors learned by Stable Diffusion and fine-tuning it on stereo image datasets to adapt it to the task of stereo generation. To improve stereo consistency and text-to-image alignment, we further tune the model using prompt alignment and our proposed stereo consistency reward functions. Comprehensive experiments demonstrate the superiority of our approach in generating high-quality stereo images across diverse scenarios, outperforming existing methods. 

\end{abstract}    
\section{Introduction}
\label{sec:intro}

With the rise in popularity of VR headsets (e.g., Meta Quest) and light field displays (e.g., Lume Pad), generating suitable content for these devices is becoming increasingly important. Although powerful diffusion models, such as Stable Diffusion~\cite{ho2020denoising, rombach2022high}, allow the average user to produce creative images from text prompts, generating stereo images remains a major challenge. 

One potential approach for generating stereo images is to first produce a single image using an existing diffusion model and then apply a single-image view synthesis method~\cite{srinivasan2019pushing, tucker2020single, 3d_photography_2020, wiles2020synsin, rockwell2021pixelsynth, koh2023simple} to reconstruct the other view. Most of these techniques~\cite{3d_photography_2020, wiles2020synsin, rockwell2021pixelsynth, koh2023simple}, however, generate novel views by warping the input image using monocular depth and inpainting the occluded regions. While these methods produce reasonable results with a small baseline, their results for larger baselines---this paper's focus---often contain objectionable artifacts. Specifically, depth-based warping often produces incorrect parallax effect for objects with continuous varying depth (see Fig.~\ref{fig:teaser}). Additionally, these methods usually reconstruct the occluded regions in a plausible but contextually inaccurate manner.

Recently, Wang et al.~\cite{wang2024stereodiffusion} tackle the problem of stereo image generation using a pre-trained Stable Diffusion model. Specifically, they follow the pipeline of previously mentioned methods, reconstructing stereo images through depth-based warping in the latent space of the diffusion model. As a result, they inherit the limitations of single-image view synthesis techniques. Furthermore, due to operating in the latent space, their warping is not pixel-perfect.

In this paper, we propose a novel diffusion-based approach for generating stereo image pairs from text prompts. Since stereo image datasets with large baselines are scarce, we supplement the existing data~\cite{tosi2023nerfstereo} by creating additional data using the multi-view MVImgNet~\cite{yu2023mvimgnet} dataset. Specifically, for each scene, we reconstruct it in 3D by optimizing a 3D Gaussian Splatting representation~\cite{3dgaussian2023} on the input images. We then rendering several stereo pairs from the optimized representation for each scene and include them as training data.

Training a diffusion model from scratch is challenging due to the limited number of scenes in our dataset; the model can easily overfit and fail to generalize well. To address this issue, we leverage the strong priors of the pre-trained Stable Diffusion~\cite{ho2020denoising, rombach2022high} and fine-tune it on our data, adapting it to the stereo generation task while retaining its generalization capabilities. However, the diffusion model outputs a single RGB image, while we are dealing with stereo image pairs. Therefore, we propose to vertically stack the left and right images to form a single RGB image, matching the output format of the diffusion model.  

This fine-tuning process adapts the diffusion model to produce stereo images, but the tuned model suffers from two issues: \textbf{1)} the generated stereo pairs are often not geometrically consistent, as consistency is not enforced during the initial fine-tuning; and \textbf{2)} while the tuned model can generate stereo images for test prompts, the generated content is often not fully aligned with the text. To address these issues, we propose using the approach by Prabhudesai et al.~\cite{alignprop2023aligning} (AlignProp), which enables fine-tuning of diffusion models according to arbitrary but differentiable reward functions. Specifically, we introduce a stereo consistency reward function to improve geometric consistency, and use human preference score v2 (HPSv2)~\cite{wu2023hpsv2} to enhance text-to-image alignment.

Experimental results demonstrate that our approach produces consistent, high-quality stereo images that outperform existing methods. In summary, our paper makes the following key contributions:

\begin{itemize}

    \item We propose fine-tuning Stable Diffusion on stereo images, adapting the model to generate stereo pairs while retaining its generalization capabilities.

    \item To improve geometric consistency and text-to-image alignment, we further tune the model using prompt alignment and our proposed stereo consistency rewards.

    \item We demonstrate that our approach produces stereo images with improved consistency and quality compared to existing methods.

\end{itemize}

\section{Related Work}
\label{sec:relatedwork}
\vspace{-0.05in}
In this section, we review closely related work, focusing on 3D generation and single-image novel view synthesis. Additionally, we discuss approaches that fine-tune diffusion models based on specific reward functions, as we leverage this technique to enhance our stereo diffusion model.

\subsection{3D Generation}

With recent advances in generative methods and 3D scene representations, such as neural radiance fields (NeRF)~\cite{mildenhall2020nerf} and 3D Gaussian Splatting~\cite{3dgaussian2023}, there has been growing interest in 3D generation. One group of methods~\cite{Niemeyer2020GIRAFFE, schwarz2020graf, EG3D2021Chan, gu2021stylenerf, gao2022get3d} integrates NeRF into generative adversarial networks (GANs)~\cite{goodfellow2014generative} to synthesize 3D content. While these methods produce high-quality results, they are limited to generating single objects.

Another category of methods leverages powerful 2D diffusion models~\cite{ho2020denoising,rombach2022high} as priors to reconstruct 3D scenes or objects. Specifically, DreamFusion~\cite{poole2022dreamfusion} and its follow-up works~\cite{melas2023realfusion, shi2023mvdream, tang2023dreamgaussian, wang2024prolificdreamer} use score distillation sampling (SDS) to optimize 3D representations like NeRF and 3D Gaussian Splatting. However, these methods often yield oversmoothed results due to SDS loss limitations, and their optimization process is computationally intensive.

Closer to our approach, some methods~\cite{Xie_2024_CVPR, li2024instantd} fine-tune Stable Diffusion~\cite{rombach2022high} on large synthetic 3D object datasets~\cite{deitke2023objaverse} to produce multi-view images, which are then passed to a transformer network to generate the final 3D representation. Xie et al.~\cite{Xie_2024_CVPR} further refine the diffusion model using reinforcement learning to enhance consistency across generated multiview results. However, these methods are primarily focused on generating individual objects. In contrast, we target stereo generation for general scenes.

\subsection{Single-Image Novel View Synthesis}

Given a single image, a large number of methods~\cite{tulsiani2018layer, 3d_photography_2020, tucker2020single, Li2020LF, pu2023sinmpi} synthesize novel views by estimating intermediate 3D representations, such as layered depth images (LDI)~\cite{shade1998layered} and multi-plane images (MPI)~\cite{zhou2018stereo}. However, these methods are generally limited to narrow viewpoint changes and struggle to generate images with significant deviations from the input.

A group of recent techniques~\cite{chung2023luciddreamer, yu2024wonderjourney, ouyang2023text2immersion, zhang2024text2nerf} leverage powerful diffusion models~\cite{ho2020denoising, rombach2022high} for this task. These methods progressively project images into the 3D scene using estimated monocular depth and inpaint occluded regions with diffusion inpainting. However, depth-based warping often produces incorrect parallax, particularly for objects with continuous varying depth. Additionally, while inpainting models can fill in occluded regions with plausible content, they frequently lack contextual accuracy. Moreover, Wang et al.~\cite{wang2024stereodiffusion} (StereoDiffusion) focus specifically on stereo generation using depth-based warping in the latent space of Stable Diffusion~\cite{rombach2022high}. However, in addition to the aforementioned issues, warping in the latent space also results in less precise pixel alignment.

\subsection{Tuning Diffusion with Rewards}

Reward fine-tuning has emerged as a promising approach to refining diffusion models, enabling the production of outputs that align with specific objectives. Inspired largely by reinforcement learning (RL), this approach has become central to applications requiring nuanced control over generation quality. For example, Lee et al.~\cite{lee2023aligning} apply reward-weighted regression on a curated dataset to address misalignments in factors such as object count, color consistency, and background quality. Methods like DDPO~\cite{ddpo2023training} and DPOK~\cite{dpok2024reinforcement} use policy gradients in multi-step diffusion models~\cite{fan2023optimizing}, enhancing reward outcomes for aesthetic quality, image-text alignment, and compressibility.

In contrast to these RL-based methods, some approaches~\cite{DRaFT2023directly,alignprop2023aligning} perform optimization by directly backpropagating gradients from a differentiable reward function, using gradient checkpointing~\cite{gruslys2016memory} to do so efficiently. In our work, we employ such techniques, particularly AlignProp~\cite{alignprop2023aligning}, to enhance the stereo consistency and prompt alignment of the generated results.

\begin{figure*}[t]
    \centering
    \includegraphics[width=1.0\linewidth]{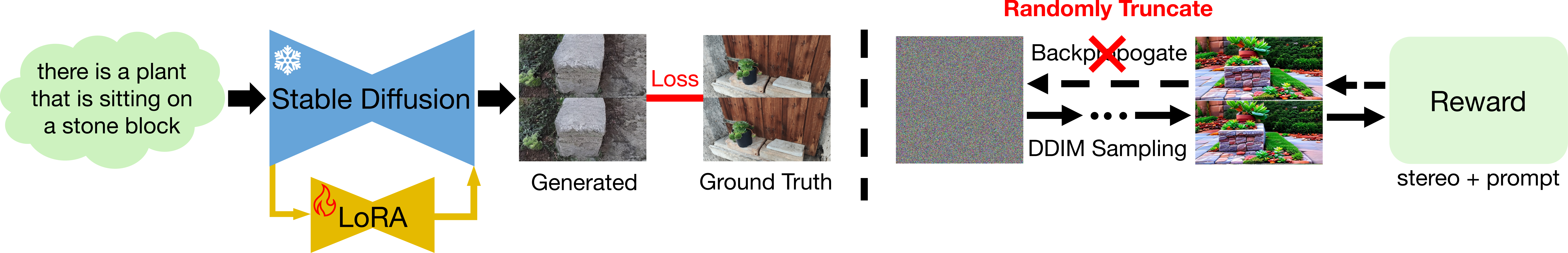} 
    \caption{We show the overview of our approach comprising two stages (left and right). In the first stage, we fine-tune the pretrained Stable Diffusion model using LoRA~\cite{hu2022lora} on our stereo image dataset, with left-right images stacked vertically. In the second stage, we further optimize our model using AlignProp~\cite{alignprop2023aligning} to enhance the stereo consistency and prompt alignment.}
    \label{fig:overview}
\vspace{-0.2in}
\end{figure*}

\section{Background}
\label{sec:background}

In this section, we provide an overview of the concepts related to our approach: diffusion models \cite{ho2020denoising, rombach2022high} and AlignProp~\cite{alignprop2023aligning}.

\subsection{Diffusion Models}

Diffusion models~\cite{ho2020denoising} are probabilistic generative models that have recently achieved state-of-the-art performance in high-quality image synthesis. The process consists of a forward and reverse diffusion phase. In the forward process, noise is gradually added to an input image $x_0$ over $T$ timesteps, producing a sequence of increasingly noisy images $x_0, \dots, x_T$, eventually leading to a noise distribution $x_T$. Specifically, given a random noise image with normal distribution $\epsilon \sim \mathcal{N}(0, \bm{I})$, the image at time $t$ is obtained by adding noise to the clean image according to $x_t = \sqrt{\bar{\alpha}_t} x_0 + \sqrt{1 - \bar{\alpha}_t} \epsilon$, where $\bar{\alpha}_t$ is derived based on the variance at each timestep.


The reverse process aims to denoise $x_T$ back to $x_0$ using a learned denoising function $\epsilon_\theta$ that takes the image at the current step $x_t$, and often a text prompt $c$, to estimate the noise $\hat{\epsilon}$, i.e., $\hat{\epsilon} = \epsilon_\theta(x_t, c, t)$. Given a set of images and corresponding text prompts $\mathcal{T} = \{(x_0^i, c^i)\}_{i=1}^{N}$, the model is trained by optimizing the following objective:

\vspace{-0.1in}
\begin{equation}
\label{eq:diffobj}
    \mathcal{L}_d = \frac{1}{N} \sum_{(x_0^i, c^i) \in \mathcal{T}} \Vert \epsilon_\theta(\sqrt{\bar{\alpha}_t} x^i_0 + \sqrt{1 - \bar{\alpha}_t} \epsilon, c^i, t) - \epsilon \Vert^2.
\end{equation}
\vspace{-0.2in}

During inference, $x_T$ is initialized with Gaussian noise, and the trained network $\epsilon_\theta$ is used to progressively denoise it, ultimately producing a clean image $x_0$. Since both training and inference of diffusion models are computationally expensive, latent diffusion models (LDM)~\cite{rombach2022high} propose performing the diffusion process in the latent space of a variational autoencoder (VAE)~\cite{kingma2013vae}, significantly reducing computational load. In our work, we utilize an LDM, specifically Stable Diffusion, and adapt it to stereo generation task.




\subsection{AlignProp}

The goal of this approach is to fine-tune a pre-trained diffusion model to produce results aligned with a specific reward. Unlike the objective in Eq.~\ref{eq:diffobj}, which operates on a single denoising step, AlignProp~\cite{alignprop2023aligning} maximizes the reward based on the model’s output after multiple denoising steps. Specifically, given a dataset of training text prompts $\mathcal{C} = \{c^i\}_{i = 1}^M$, AlignProp optimizes the diffusion model’s parameters to maximize the following objective function:

\vspace{-0.05in}
\begin{equation}
    \mathcal{L}_a = - \frac{1}{M} \sum_{c^i \in \mathcal{C}} R(\pi_\theta(x_T, c^i)),
\end{equation}
\vspace{-0.1in}

\noindent where $c^i$ represents a training prompt and $R$ is the reward function, which might, for example, measure aesthetic quality or compressibility of the generated images $x_0$. Moreover, $\pi_\theta$ encapsulating the iterative denoising process into a single function, i.e., $x_0 = \pi_\theta(x_T, c)$.

Optimizing this objective by fully backpropagating through all denoising steps, however, leads to mode collapse. To address this issue, AlignProp proposes truncating gradient backpropagation at a random denoising step. This adjustment enables the optimization process to adapt the network according to the reward while avoiding mode collapse. In our work, we use AlignProp to enhance stereo consistency and prompt alignment in the generated results.



\section{Methodology}
\label{sec:algorithm}


The goal of our work is to train a diffusion model that generates consistent stereo image pairs with a large baseline from text prompts. Our approach, dubbed Text2Stereo, adapts the pre-trained Stable Diffusion~\cite{rombach2022high} model to the stereo generation task. In the following sections, we first discuss our process for obtaining a dataset of stereo images and their corresponding text prompts (Sec.~\ref{ssec:dataset}). We then describe our fine-tuning process, which consists of two stages: stereo adaptation (Sec.~\ref{ssec:stereo_adaptation}) and fine-tuning for consistency enhancement (Sec.~\ref{ssec:consistency_enhancement}). Overview of our approach is illustrated in Fig.~\ref{fig:overview}.



\begin{figure}
    \centering
    \includegraphics[width=1.0\linewidth]{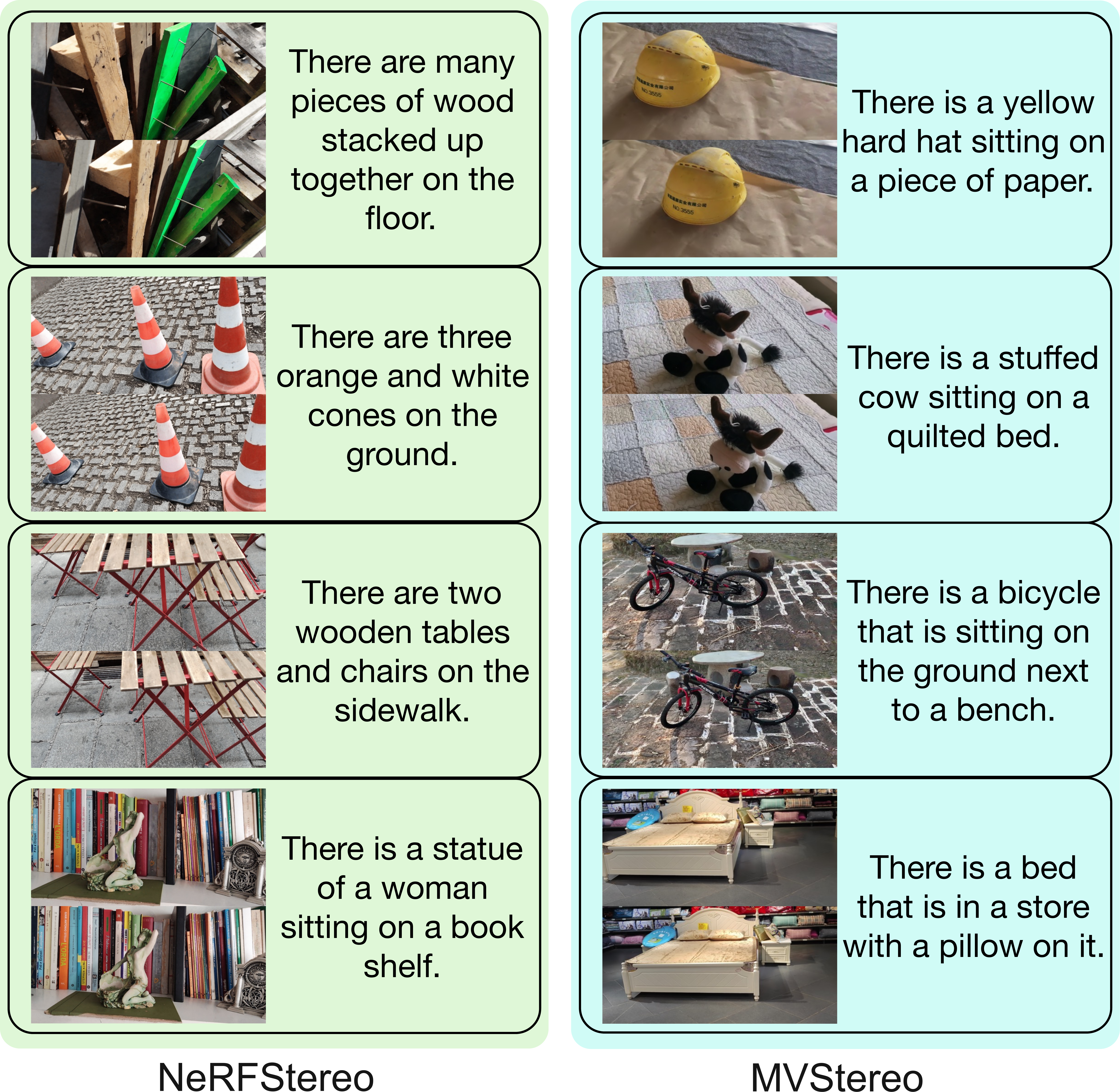} 
    \caption{We present example images from our dataset. In addition to the existing stereo dataset from NeRFStereo~\cite{tosi2023nerfstereo}, we introduce a newly generated dataset, MVStereo. This dataset is created by first reconstructing a 3D Gaussian splatting~\cite{3dgaussian2023, zhu2024fsgs} representation from multiview images obtained from MVImgNet~\cite{yu2023mvimgnet}, followed by sampling stereo images from the reconstructed model.}
    \label{fig:datasets}
\vspace{-0.2in}
\end{figure}

\subsection{Dataset Preparation}
\label{ssec:dataset}

Since the goal of our work is to generate stereo images with a large baseline, we need a dataset of such stereo image pairs for training. Unfortunately, most existing stereo datasets are either captured with stereo cameras that have a small baseline~\cite{Holopix} or are synthetic~\cite{SceneFlow}, making them unsuitable for our task. A notable exception is the dataset by NeRFStereo~\cite{tosi2023nerfstereo}, which contains 270 scenes, each with 100 large baseline stereo images (total 27,000).

In our work, we supplement the NeRFStereo dataset by creating our own stereo image dataset, coined MVStereo, using the multi-view MVImgNet dataset~\cite{yu2023mvimgnet}. Specifically, we select a subset of 234 scenes from MVImgNet~\cite{yu2023mvimgnet} and reconstruct them in 3D by optimizing a 3DGS representation~\cite{3dgaussian2023}, utilizing the approximately 30 images available for each scene. Although the 3DGS optimization~\cite{3dgaussian2023} yielded reasonable results, we found that the method proposed by Zhu et al.~\cite{zhu2024fsgs} produced better outcomes with fewer floaters and less blurriness, which is why we adopt their approach to reconstruct the scenes. 

Once the 3DGS representation is obtained, we render 7373 stereo pairs by setting up stereo cameras at various views. We carefully position the cameras around the input scene to ensure the rendered images do not contain floaters or blurry content. Together, we use 30982 stereo images across 458 scenes; 234 from MVStereo and a subset of 224 scenes from NeRFStereo, both encompassing diverse indoor and outdoor environments. Since our goal is text-based stereo generation, we also require corresponding text prompts for each stereo image in our dataset, which we obtain using the BLIP captioning model~\cite{BlipImageCaptioning}. Some examples of stereo images and their corresponding captions from both NeRFStereo and our MVStereo are shown in Fig.~\ref{fig:datasets}.

\subsection{Stereo Adaptation}
\label{ssec:stereo_adaptation}

Given a dataset of stereo images and their corresponding text prompts, $\{x_l^i, x_r^i, c^i \}_{i=1}^N$, we aim to train a stereo generator. However, due to the relatively small size of our dataset, training a diffusion model from scratch is impractical. Even with a larger dataset, ensuring generalization to diverse text prompts would be challenging. To address this, we leverage the strong prior knowledge in a pre-trained diffusion model, specifically Stable Diffusion~\cite{rombach2022high}, and adapt it for the stereo generation task.

The primary challenge here is that Stable Diffusion is designed to produce a single RGB image, whereas our objective is to generate stereo image pairs. Inspired by Instant3D~\cite{li2024instantd}, we propose stacking our stereo image pairs, $x_l^i$ and $x_r^i \in \mathbb{R}^{256\times 512 \times 3}$, vertically to form a single RGB image, $x^i \in \mathbb{R}^{512\times 512 \times 3}$ (see Fig.~\ref{fig:overview}). This stacked representation, along with the corresponding text prompts, forms our training dataset, $\mathcal{T} = \{x^i, c^i\}_{i=1}^N$, which we use to fine-tune Stable Diffusion according to the objective in Eq.~\ref{eq:diffobj}. Note that we do not provide camera pose information as the input to the network. However, since the training data is mainly composed of stereo images with a large baseline, our trained model has the ability to produce such stereo pairs.

Fine-tuning Stable Diffusion by directly optimizing its parameters, $\theta$, can lead to overfitting due to the small size of our dataset. To address this, we employ Low-Rank Adaptation (LoRA)~\cite{hu2022lora}, which mitigates overfitting by freezing the diffusion parameters $\theta$ and modulating them through a trainable layer with significantly fewer parameters. Specifically, each linear layer, initially represented as $h = Wx$ with $W \in \mathbb{R}^{d\times d}$, is modified to $h = Wx + BAx$, where $A \in \mathbb{R}^{d \times r}$ and $B \in \mathbb{R}^{r \times d}$, with $r \ll d$. By freezing $W$ and updating only $A$ and $B$ in each layer of Stable Diffusion, we adapt the model for stereo generation while preserving its ability to generalize.





\subsection{Fine-Tuning for Consistency Enhancement}
\label{ssec:consistency_enhancement}

Our fine-tuned model, as shown in Fig.~\ref{fig:ablation} (Base), produces vertically stacked stereo images for unseen text prompts. However, the generated results exhibit two issues. First, while the content in the left and right images shifts with the camera's perspective, there are often deformations and inconsistencies between objects in the two views. This issue primarily arises because our initial fine-tuning lacks a mechanism to enforce consistency between the left and right images. Second, we observe that after fine-tuning, the diffusion model generates images that, in some cases, are not fully consistent with the text prompt.

To address these issues, we propose further fine-tuning the model to enhance its stereo and prompt consistency using AlignProp~\cite{alignprop2023aligning}. The main challenge here is designing an appropriate reward function $R$ that measures the stereo and prompt consistency of the generated images.
To this end, we propose a reward function consisting of three terms as follows:

\vspace{-0.1in}
\begin{equation}
\label{eq:consreward}
    R = \alpha R_s + \beta R_p + \gamma R_c
\end{equation}

\noindent where $R_s$, $R_p$, and $R_c$ refer to the stereo consistency, prompt consistency, and convergence rewards, described below. Moreover, $\alpha = 0.25$, $\beta = 0.75$, and $\gamma = 0.25$ define each term's weight.


\begin{figure}
    \centering
    \includegraphics[width=1.0\linewidth]{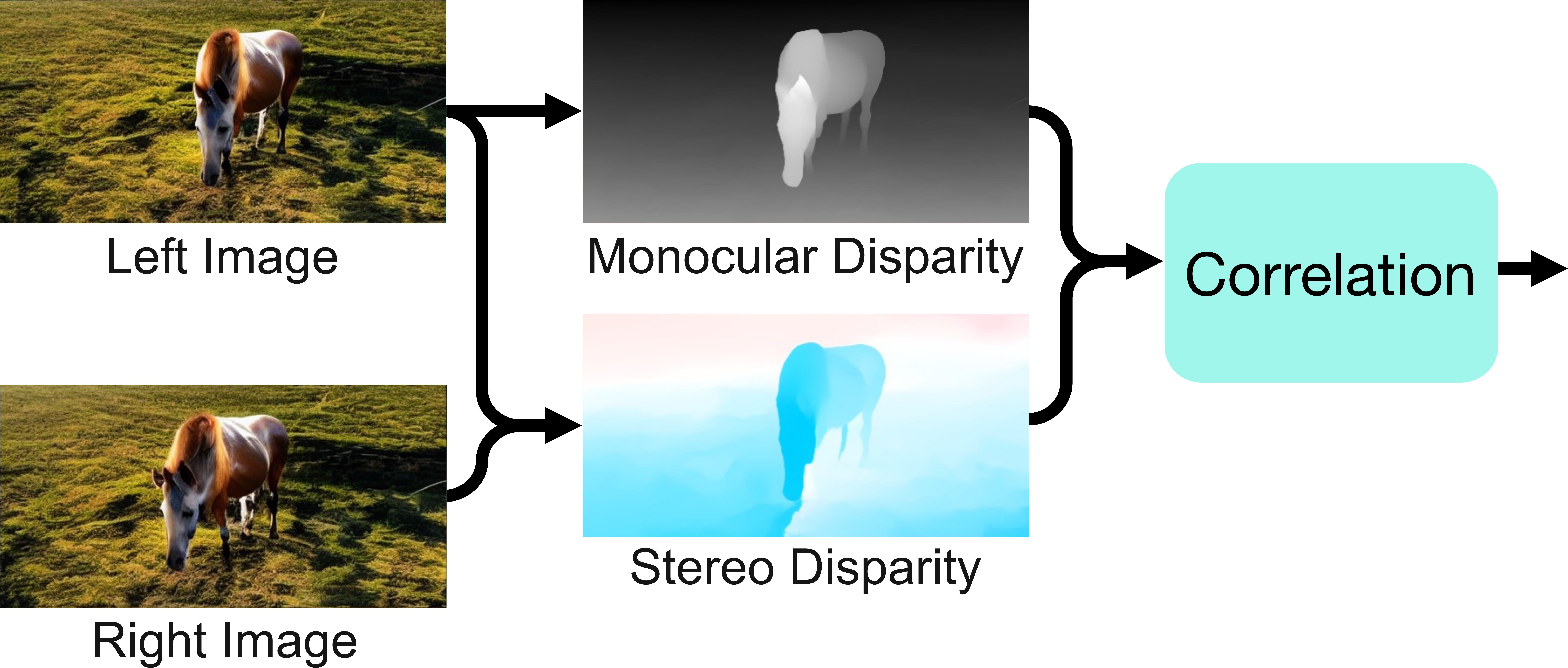} 
    \vspace{-0.2in}
    \caption{Given a stereo image pair, we estimate the stereo disparity using both images and monocular disparity using only the left image. The correlation between the two maps will then serve as our stereo consistency reward function.}
    \label{fig:stereo_reward}
\vspace{-0.15in}
\end{figure}

\vspace{-0.1in}
\paragraph{Stereo Consistency:} Since there is currently no well-established mechanism for checking the stereo consistency between generated stereo pairs, we need to design our own metric. Our key idea is that for stereo images to be consistent, the stereo and monocular disparities should align. This ensures that the model avoids trivial solutions, such as duplicating content, where the stereo disparity is zero, but the monocular disparity still reflects the correct depth. To achieve this, we estimate the stereo disparity as $d^s = \Phi(x_l, x_r)$ and the monocular disparity as $d^m = \Psi(x_l)$ and measure their similarity. Since the monocular disparity is relative, comparing these two disparities directly using pixel-wise metrics, such as $L_2$, is not effective. Therefore, we propose measuring their similarity using Pearson correlation, as follows:

\vspace{-0.15in}
\begin{equation}
    R_s = \frac{\sum_p (d^m(p) - \bar{d^m})(d^s(p) - \bar{d^s})}{\sqrt{\sum_p (d^m(p) - \bar{d^m})^2 \sum_p (d^s(p) - \bar{d^s})^2}},
\end{equation}
\vspace{-0.1in}

\noindent where $\bar{d^m}$ and $\bar{d^s}$ are the average monocular and stereo disparities over all pixel coordinates $p$. In our implementation, we use DepthAnythingV2~\cite{yang2024depthv2} to estimate monocular disparity. For stereo disparity, we initially experimented with CREStereo~\cite{li2022crestereo}, but found it to be sensitive to imperfections in the stereo pairs. Therefore, we use SEA-RAFT~\cite{wang2025sea} to estimate the optical flow between the two images, and then use the $x$-coordinate as the disparity. We illustrate our stereo consistency reward in Fig.~\ref{fig:stereo_reward}.

\vspace{-0.1in}
\paragraph{Prompt Consistency:} We use the human preference score v2~\cite{wu2023hpsv2}, which trains a CLIP model~\cite{clip} on a large annotated dataset. This score reliably measures the consistency between the text prompt and the generated images, and we adopt it as our prompt consistency metric, $R_p$.

\vspace{-0.1in}

\paragraph{Convergence:} Stereo images captured with cameras with parallel optical axis have convergence at infinity, i.e., the objects at infinite depth will have zero disparity. Through the finetuning, however, the diffusion model may generate stereo images with convergence at the middle of the scene, i.e., objects further away from the convergence will have negative disparity. To avoid this issue, we introduce the following reward to penalize the negative disparities:

\vspace{-0.05in}
\begin{equation}
    R_c =  - \frac{\Vert \max(- d^s(p), 0) \Vert_1}{\max(\max(- d^s), 1)}.
\end{equation}
\vspace{-0.1in}

Here, the denominator is a normalization factor that prevents large negative disparities (greater than 1) from causing a spike in the loss.



\section{Results}
\label{sec:results}

\begin{figure*}
    \centering
    \includegraphics[width=1.0\linewidth]{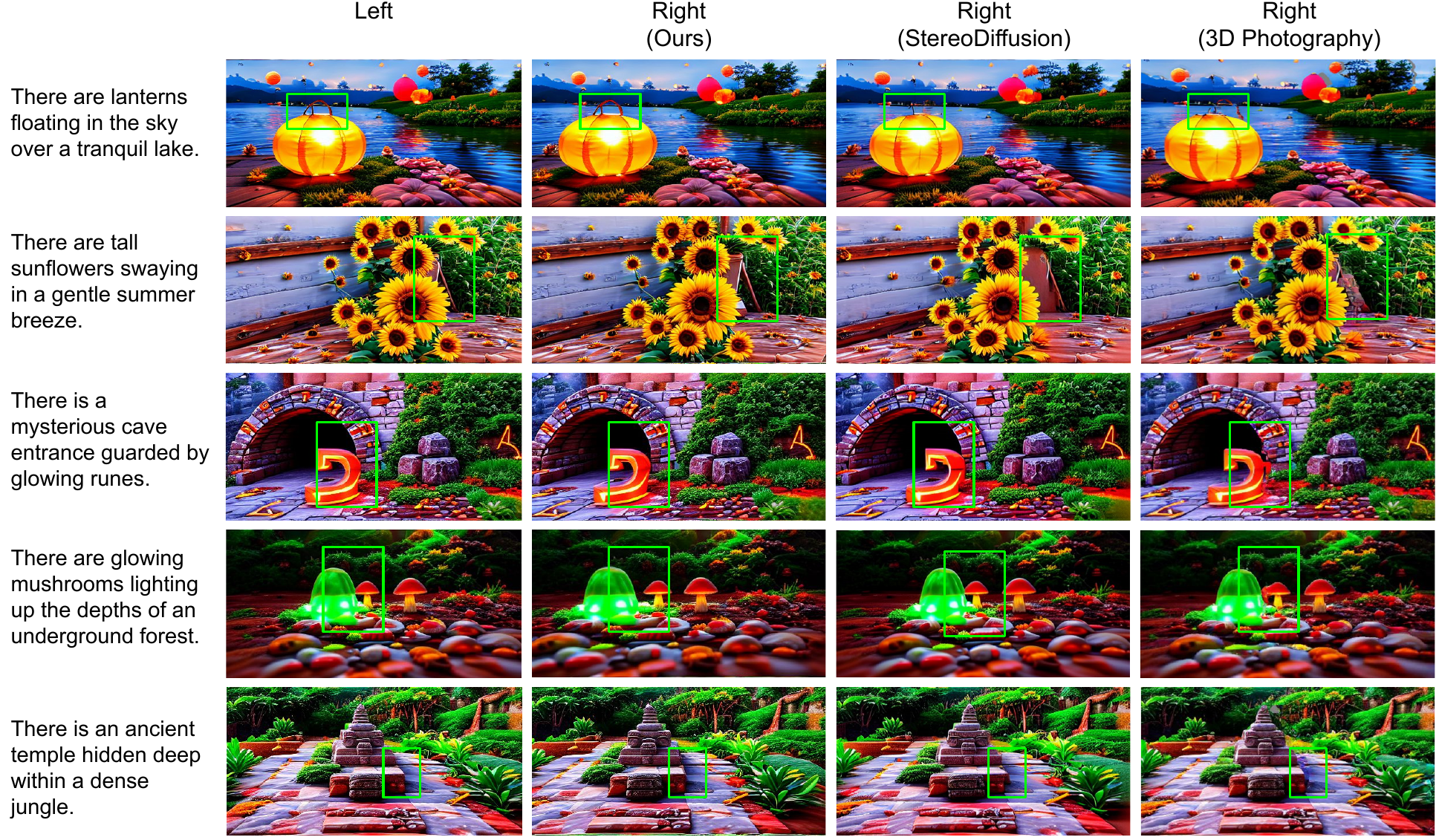} 
    \vspace{-0.2in}
    \caption{Qualitative comparison of our approach against StereoDiffusion~\cite{wang2024stereodiffusion} and 3D Photography~\cite{3d_photography_2020} on five test prompts. Given a text prompt, we generate a stereo pair and use the left image as input for the other methods to reconstruct the right image. StereoDiffusion, which performs warping in the latent space, often distorts objects (top two rows) or fails to position them correctly (bottom three rows). For example, note that StereoDiffusion does not produce the gap between the mushrooms in the fourth example. 3D Photography struggles with depth inaccuracies (e.g., thin structures in the top row) and fails to reconstruct occluded areas (bottom four rows). In contrast, our approach produces consistent, high-quality results with wide baselines.}
    \label{fig:visual_comparison}
\vspace{-0.2in}
\end{figure*}

In this section, we first describe the implementation details. We then show comparisons against state-of-the-art methods and demonstrate the impact of various components of our approach.

\subsection{Implementation Details}

We implement our method in PyTorch and use the pretrained Stable Diffusion v1.5 as our base model. Additionally, we utilize LoRA with rank 4, and inject it into every U-Net cross-attention layer. In the initial training phase, we optimize the model with a cosine learning rate scheduler starting at 1e-4, employing a batch size of 4 with gradient accumulation over 4 steps for a total of 4000 iterations, which took roughly 6 hours on a single A100 GPU. Subsequently, for optimization using consistency rewards, the model undergoes further fine-tuning for 300 iterations with a batch size of 100 prompts per step utilizing 4 A100 GPUs for a day. 



\subsection{Comparisons}

We demonstrate the effectiveness of our approach by providing comparisons against StereoDiffusion~\cite{wang2024stereodiffusion} and 3D Photography~\cite{3d_photography_2020}. Specifically, in each case, we generate the stereo images given a text prompt with our approach and use the left image as the input to the other techniques for reconstructing the right image. As such, we only compare the reconstructed right images. 3D Photography reconstruct the novel image through depth-based warping and inpainting, while StereoDiffusion, performs the warping in the latent space of a diffusion model.

Figure~\ref{fig:visual_comparison} shows comparisons against the other approaches on five test prompts. Since the warping in the latent domain is not precise, StereoDiffusion either distort the objects (top two scenes) or is unable to move various objects to the appropriate location (last three scenes). Similarly, 3D Photography struggles in cases where the depth is inaccurate (e.g., the thin structure in the top scene) and is unable to properly reconstruct the occluded areas (bottom four scenes). In contrast, our approach produces consistent high-quality results with large baselines.


\begin{figure*}
    \centering
    \includegraphics[width=1.0\linewidth]{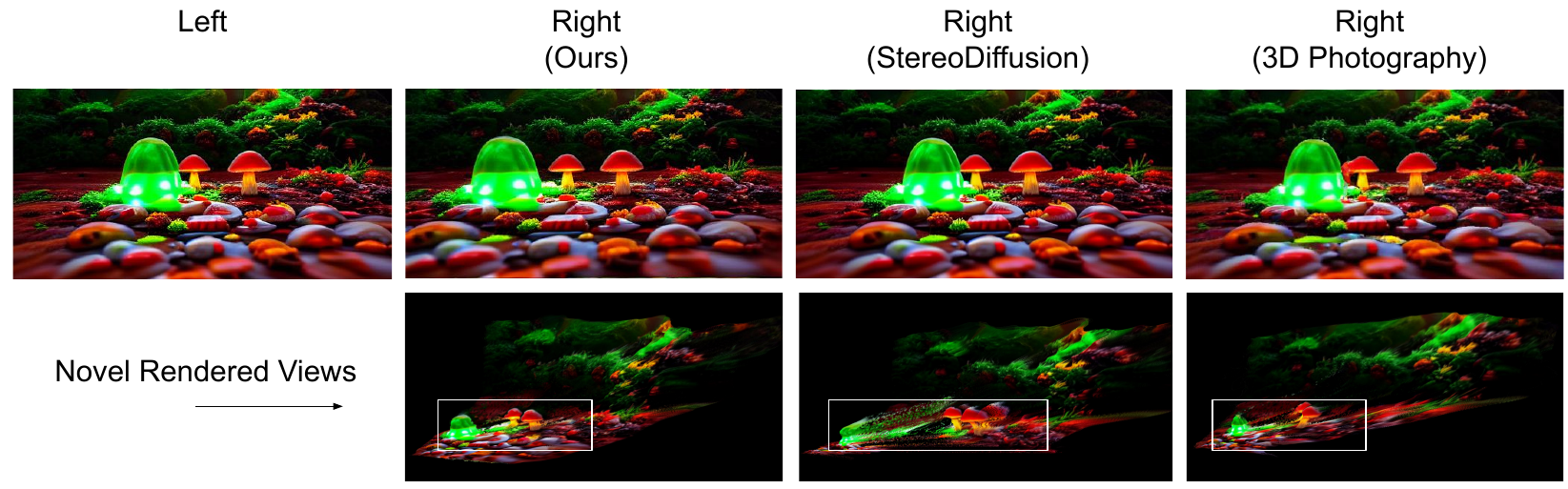} 
    \caption{Evaluation of stereo consistency using Splatt3R~\cite{smart2024splatt3r}. Given the text prompt: ``There are glowing mushrooms lighting up the depths of an underground forest,'' we generate a stereo pair with each method and use it as input to Splatt3R to obtain a 3D Gaussian splatting (3DGS)\cite{3dgaussian2023} representation. The 3DGS model is then rendered from a novel viewpoint. StereoDiffusion\cite{wang2024stereodiffusion} and 3D Photography~\cite{3d_photography_2020} produce stereo images with inconsistencies, leading to artifacts in the rendered view. In contrast, our approach generates more consistent stereo pairs, resulting in a high-quality 3DGS representation with clear object boundaries.}
    \label{fig:splatt3r}
\end{figure*}

\begin{figure*}
    \centering
    \includegraphics[width=1.0\linewidth]{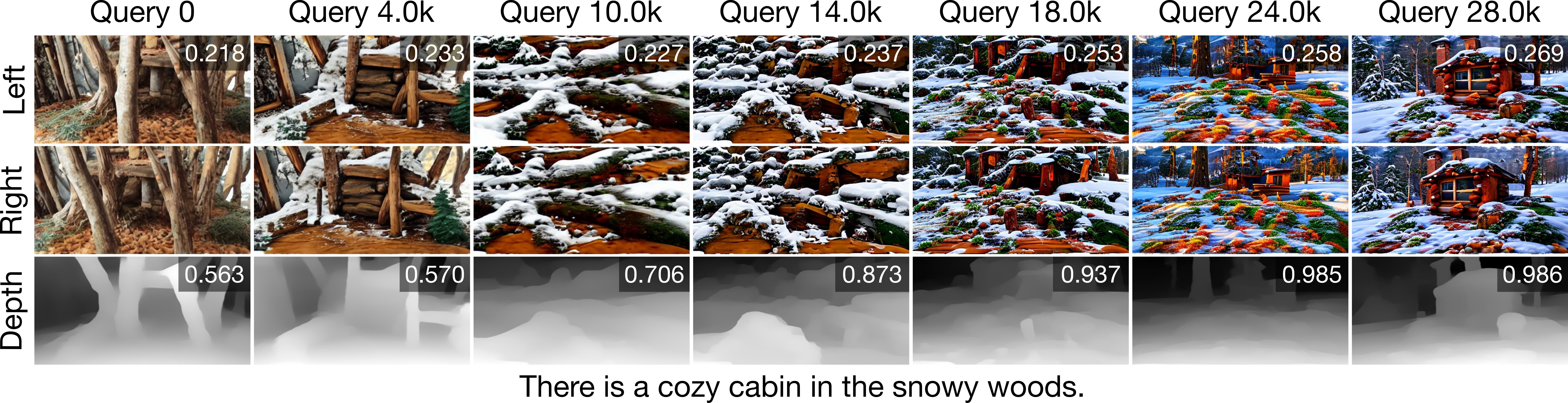} 
    \caption{We demonstrate the results through various stages of the training process. The numbers in the first and third rows correspond to the prompt alignment score and stereo consistency score, respectively. As seen, both stereo consistency and prompt alignment exhibit improvements throughout the reward optimization process.}
    \label{fig:reward_process}
\vspace{-0.1in}
\end{figure*}

We further evaluate the consistency of generated results in Figure~\ref{fig:splatt3r}. Specifically, we use the pair of images generated by each method as the input to Splatt3R~\cite{smart2024splatt3r} to obtain the corresponding 3D Gaussian splatting (3DGS)~\cite{3dgaussian2023} representation. We then render the 3DGS representation from a novel view and compare the renderings. The key idea is that if the stereo images are consistent, Splatt3R will produce a high-quality 3DGS representation and thus the rendered images will be of high quality. As shown in Figure~\cite{smart2024splatt3r}, the novel view images for both StereoDiffusion and 3D Photography contain distracting artifacts. In contrast, the rendering by our approach has clear object boundaries, demonstrating the consistency of our generated left and right images.

\subsection{Analysis}

 We begin by showing the progressive improvement of stereo consistency and prompt alignment with the optimization of our consistency rewards (Eq.~\ref{eq:consreward}) in Fig.~\ref{fig:reward_process}. The values presented in the first and third rows represent the prompt and stereo consistency scores, respectively. Each column is obtained by generating the results of the network after certain number of reward queries. Here the reward query refers to the number of times the reward function is evaluated during the fine-tuning process, e.g., one iteration with a batch size of 100 leads to 100 reward queries. As seen both the prompt and stereo consistency of the results improve during the optimization (despite some fluctuations). Particularly, the stereo depth maps (third row) improve significantly through the fine-tuning process.



\begin{figure}
    \centering
    \includegraphics[width=1.0\linewidth]{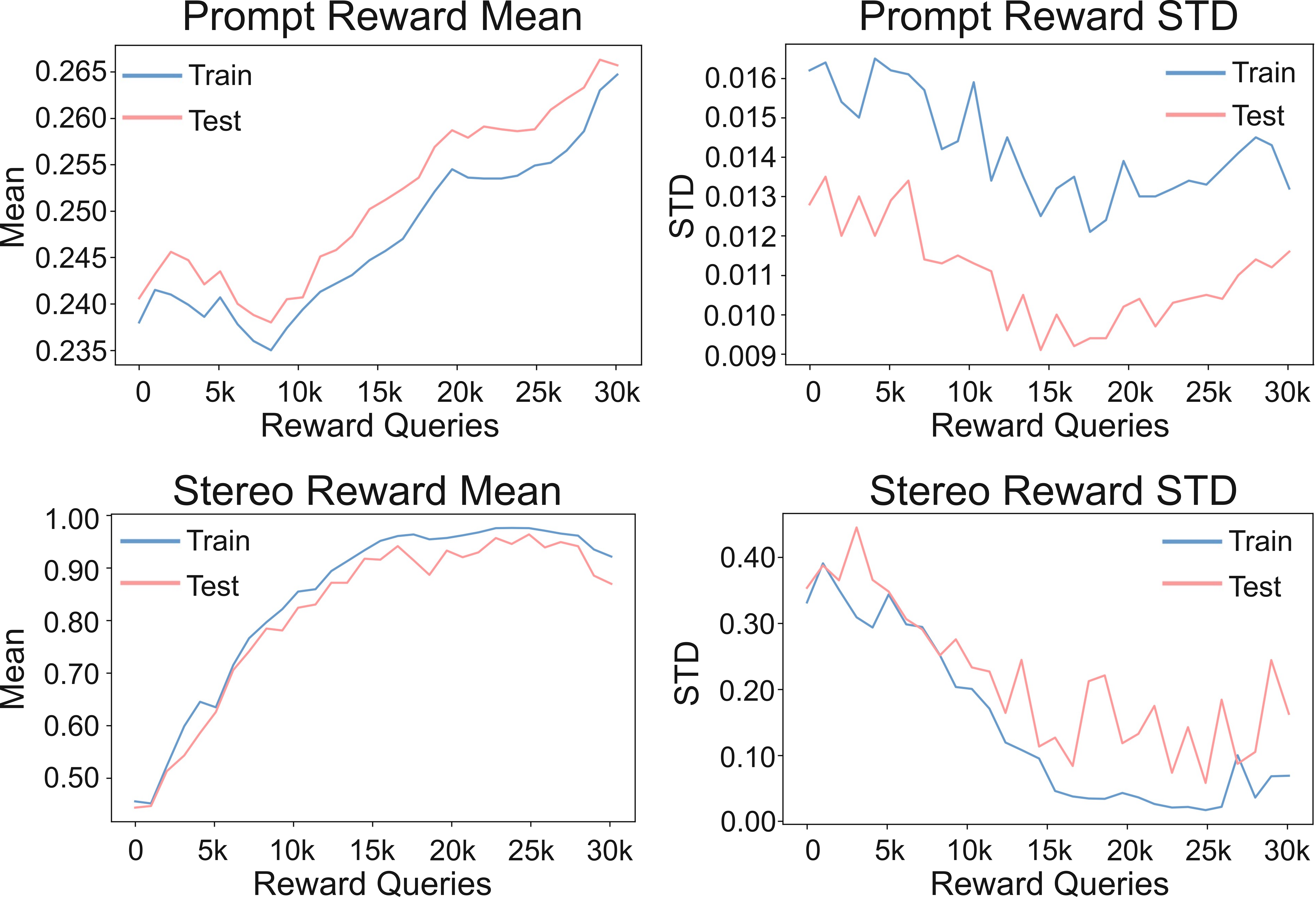} 
    \caption{We show the rewards averaged over 100 training and test prompts during the optimization process. The mean values of both prompt and stereo consistency rewards increase, while their standard deviations decrease, during the reward optimization process. These trends suggest an improvement in the model's ability to generate images with enhanced prompt alignment and stereo consistency. Furthermore, across all plots, both training and testing results demonstrate similar trends, indicating that optimizing for consistency objectives does not adversely affect the model’s generalization capabilities.}
    \label{fig:reward_plots}
\end{figure}

We further evaluate how the prompt and stereo consistency rewards evolve during the optimization for both training and test prompts in Figure~\ref{fig:reward_plots}. Specifically, we perform the evaluation on 100 training prompts, drawn from the training dataset, and 100 test prompts, generated by ChatGPT 4o. We ensure that the test prompts are substantially different from the training prompts. As shown in Fig.~\ref{fig:reward_plots}, during the reward optimization process, both the mean values of the prompt and stereo consistency rewards exhibit a progressive increase, while their standard deviations decrease. These trends indicate an enhancement in the model’s capability to generate images with improved prompt alignment and stereo consistency. Additionally, in all plots, the training and testing results follow similar trajectories, suggesting that optimizing for consistency objectives does not compromise the model’s generalization ability.

\begin{figure}
    \centering
    \includegraphics[width=1.0\linewidth]{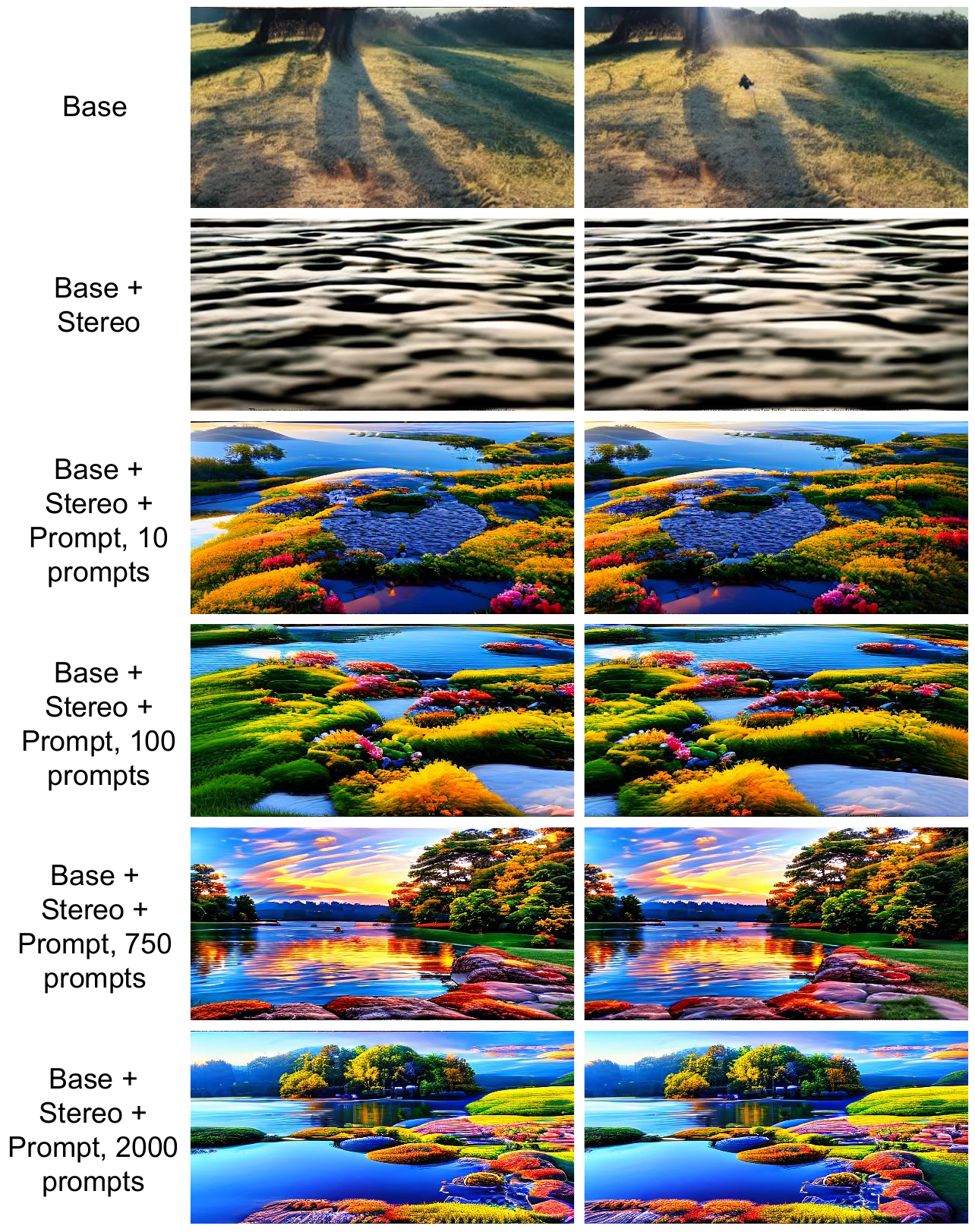} 
    \vspace{-0.2in}
    \caption{We show the impact of different component of our system visually. The test prompt is: ``There is a serene sunrise over a calm lake, promising a day filled with wonder.''}
    \label{fig:ablation}
\vspace{-0.15in}
\end{figure}

Finally, we conduct ablation studies to evaluate the impact of various components of our system both numerically (Table~\ref{tab:ablation}) and visually (Figure~\ref{fig:ablation}). As seen, the base model (without consistency tuning) produces results with poor stereo consistency and prompt alignment. Optimizing with only the stereo consistency reward (Base + Stereo) improves stereo consistency but significantly reduces prompt alignment. Optimizing with both stereo and prompt consistency rewards improves both scores. Additionally, increasing the training prompts to 750 consistently enhances the results, after which the improvement stabilizes.
Note that compared to the variant without prompt consistency (Base + Stereo), our method produces results with slightly lower stereo scores. However, as shown in Fig.~\ref{fig:ablation}, our approach produces stereo images with the best trade off between stereo and prompt consistency.


\begin{table}[t]
\centering
\caption{We quantitatively evaluate the impact of various components of our system in terms of the stereo and prompt consistency scores.}
\label{tab:ablation}
\resizebox{\columnwidth}{!}{%
\begin{tabular}{lcc}
\hline
\textbf{Condition} & \textbf{Stereo Score} & \textbf{Prompt Score} \\
\hline
Base                                         & 0.414 $\pm$ 0.327               & 0.237 $\pm$ 0.012                \\
Base + Stereo                                & \textbf{0.985  $\pm$ 0.012}              & 0.205 $\pm$ 0.010               \\
Base + Stereo + Prompt, 10 prompts           & 0.877 $\pm$ 0.192   & 0.258 $\pm$ 0.010    \\
Base + Stereo + Prompt, 100 prompts          & 0.937 $\pm$ 0.098   & 0.254 $\pm$ 0.010    \\
Base + Stereo + Prompt, 750 prompts (Ours)     & 0.949 $\pm$ 0.078   & \textbf{0.264 $\pm$ 0.011}    \\
Base + Stereo + Prompt, 2000 prompts         & 0.940 $\pm$ 0.091   & 0.263 $\pm$ 0.011    \\
\hline
\end{tabular}%
}
\end{table}

\section{Conclusion, Limitations, and Future Work}
\label{sec:conclusion}

We have presented a novel method for generating wide-baseline stereo images by adapting a pre-trained diffusion model to this task. Specifically, we first fine-tune a Stable Diffusion model on a stereo dataset to produce vertically stacked left and right images. Moreover, to improve stereo consistency and prompt alignment, we propose specific reward functions used to further tune the model. In particular, we introduce a stereo consistency reward that calculates the similarity of monocular and stereo disparities using Pearson correlation. Through experimental results, we demonstrate that our approach outperforms existing methods.

Despite producing high-quality results, our approach has a few limitations. For example, it currently does not provide a mechanism to control the baseline of the generated stereo images. In the future, it would be interesting to investigate a way to use the baseline as an input to the diffusion process to enhance controllability. Additionally, our approach can generate stereo images only from a text prompt and cannot reconstruct stereo images from a single image. One potential solution is to invert the image into our diffusion process to reconstruct the other view. We leave the investigation of this strategy to future work. Finally, we observed that the captions generated by the BLIP model are short and could sometimes be inaccurate. In the future, it would be interesting to utilize more descriptive image captioning approaches such as LLaVA~\cite{liu2023visualinstructiontuning}, combined with human verification of the captions, as they have been shown to improve the image generation quality~\cite{segalis2023pictureworththousandwords}.

\section*{Acknowledgements}
\label{sec:acknowledgements}

The project was funded by Leia Inc. (contract \#415290).
Portions of this research were conducted with the advanced
computing resources provided by Texas A\&M High Performance Research Computing.
\newpage
{
    \small
    \bibliographystyle{ieeenat_fullname}
    \bibliography{main}
}


\end{document}